\documentclass[11pt,letterpaper]{article}

\pdfoutput=1

\usepackage[T1]{fontenc}
\usepackage[utf8]{inputenc}
\usepackage{times}
\usepackage{latexsym}
\usepackage{inconsolata}

\usepackage[final]{ACL2023} 

\usepackage{microtype}

\usepackage{amsmath}
\usepackage{amssymb}
\usepackage{amsthm}

\usepackage{algorithm}
\usepackage{algpseudocode}

\usepackage{booktabs}
\usepackage{multirow}
\usepackage{graphicx} 
\usepackage{graphicx}
\usepackage{tikz}
\usepackage{tabularx} 

\usepackage{xcolor}

\usepackage{hyperref}

\usepackage{natbib}

\usepackage{xcolor}
\usepackage{array} 
\usepackage{tabularx} 
\usepackage{multirow}  
\usepackage{float}

\usepackage{listings}
\usepackage{xcolor}

\title{PatentScore: Multi-dimensional Evaluation \\of LLM-Generated Patent Claims}

\makeatletter
\ifx\pdfpagewidth\undefined\else
\fi
\makeatother

\author{
Yongmin Yoo, \quad Qiongkai Xu, \quad Longbing Cao \\
Frontier AI Research Centre, Macquarie University \\
School of Computing, FSE, Macquarie University\\
\texttt{yooyongmin91@gmail.com} \quad 
\texttt{\{qiongkai.xu, longbing.cao\}@mq.edu.au}
}

\begin{document}
\maketitle
\begin{abstract}
 High-stakes texts such as patent claims, medical records, and technical reports are structurally complex and demand a high degree of reliability and precision. While large language models (LLMs) have recently been applied to automate their generation in high-stakes domains, reliably evaluating such outputs remains a major challenge. Conventional natural language generation (NLG) metrics are effective for generic documents but fail to capture the structural and legal characteristics essential to evaluating complex high-stakes documents. To address this gap, we propose PatentScore, a multi-dimensional evaluation framework specifically designed for one of the most intricate and rigorous domains, patent claims. PatentScore integrates hierarchical decomposition of claim elements, validation patterns grounded in legal and technical standards, and scoring across structural, semantic, and legal dimensions. In experiments on our dataset which consists of 400 Claim1, PatentScore achieved the highest correlation with expert annotations ($r = 0.819$), significantly outperforming widely used NLG metrics. This work establishes a new standard for evaluating LLM-generated patent claims, providing a solid foundation for research on patent generation and validation.
\end{abstract}

\section{Introduction}

Large language models (LLMs) have demonstrated impressive capabilities in automated text generation across a wide range of natural language processing tasks. However, existing evaluation frameworks remain insufficient for rigorously and systematically assessing their outputs in high-stakes domains. These domains, such as patent claims, legal contracts, and technical reports, involve deeply hierarchical and interdependent structures that cannot be accurately evaluated using surface-level metrics such as sentence-level fluency or shallow contextual similarity. Current evaluation metrics often fail to capture critical domain-specific nuances, including legal constraints, structural dependencies, technical precision, and implicit reasoning, which significantly limit their effectiveness in these contexts. This limitation is especially consequential because it poses risks to both legal validity and practical utility of the generated documents, thereby hindering their adoption in real-world, risk-sensitive applications~\citep{holzenberger2023legalbench}.

\begin{figure}[t!]
    \centering
    \includegraphics[width=0.95\linewidth, keepaspectratio]{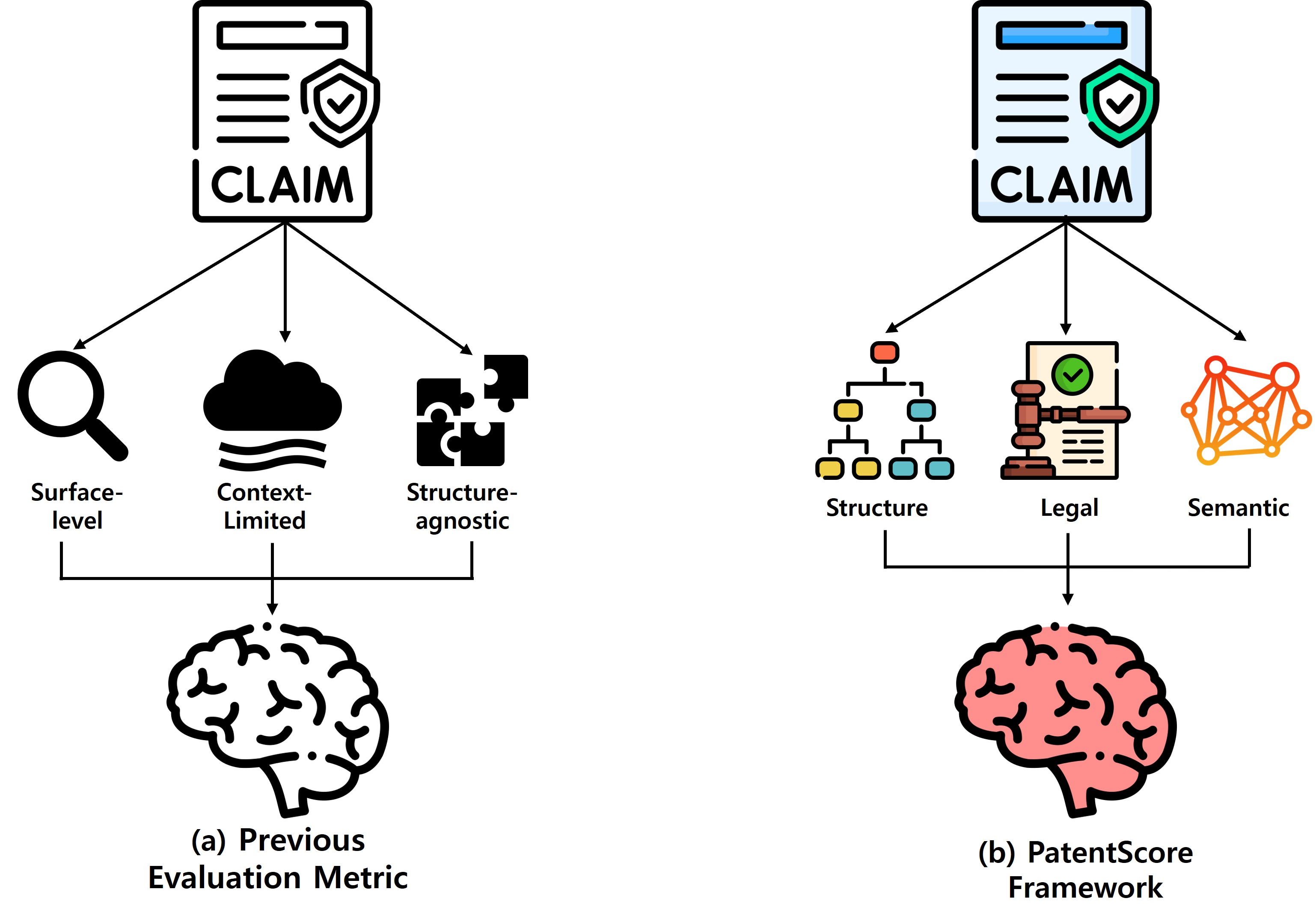}
    \caption{A comparison of PatentScore and standard NLG metrics in evaluating structural and legal accuracy of patent claims.}
    \label{fig:comparison_metrics}
\end{figure}

Among high-stakes legal and technical documents, patent claims pose particularly significant challenges for evaluation. Because they function as both technical descriptions and legally binding texts, patent claims require precise assessment of structural elements such as antecedent consistency and claim dependency. Existing natural language generation (NLG) evaluation methods struggle in domains where both linguistic precision and legal accountability are critical. Patent claims exemplify these characteristics, as they define the scope of rights granted to an invention. The first claim, in particular, plays a central role in shaping the structure and coverage of the entire claim set. This pivotal role entails substantial socio-economic implications, as it determines the legal and commercial value of the invention~\citep{merges1997patent, bessen2009patent, surden2019artificial}.

With the rapid advancement of LLMs, recent studies have explored their potential to generate and evaluate precision-critical legal texts. However, patent claims remain a particularly challenging domain due to their intricate structural and legal constraints~\citep{aristodemou2018, lee2020}. Traditional NLG metrics, such as BLEU~\citep{papineni2002bleu}, ROUGE~\citep{lin2004rouge}, and BERTScore~\citep{zhang2019bertscore}, primarily focus on lexical overlap or contextual similarity, failing to address the structural and legal nuances required for patent claim evaluation. 

More recent LLM-based evaluation approaches, such as LLM-as-a-Judge~\citep{zheng2023judging}, rely on general-purpose models to assess fluency and semantic coherence. However, they lack the domain-specific knowledge required to evaluate structural constraints, including claim dependency and legal clarity, that are critical to patent claims. These shortcomings limit the applicability of such metrics in high-stakes legal contexts where precision is paramount~\citep{zuo2024patenteval}. For instance, inconsistent antecedent references can render a patent legally ambiguous, while punctuation errors may distort its scope of protection~\citep{merges1997patent}.

To address these limitations, we propose PatentScore, a multi-dimensional evaluation framework for patent claims. This framework evaluates structural and legal elements such as antecedent consistency, claim dependency, and legal clarity. Our approach decomposes claims into constituent parts and applies targeted metrics, enabling reliable and consistent assessment of LLM-generated claims. This systematic evaluation significantly enhances the quality of claim evaluation compared to existing metrics. Our objective is to develop a domain-specific metric for patent claims while establishing a generalizable, structure-aware evaluation framework applicable to high-precision documents, such as contract clauses, policy reports, and regulatory compliance in clinical documentation. 

The key contributions of PatentScore are summarized as follows:
\begin{itemize}
    \item \textbf{Claim-Structured Evaluation Framework.} We introduce an architecture that decomposes a claim into structural components and quantifies each, covering claim structure, element linkage, antecedent basis, and claim dependency per official examination guidelines.
    \item \textbf{Expert-Aligned Validation.} We compare PatentScore’s evaluation of LLM-generated claims against expert ratings, achieving a strong correlation ($r = 0.819$) and outperforming standard NLG metrics in capturing structural and legal precision.
    \item \textbf{Open Evaluation Benchmark.} We release 400 LLM-generated claims with expert annotations, creating the first public benchmark for evaluating LLM-generated patent texts.
\end{itemize}

Although our experiments focus on patent claims, PatentScore is model-agnostic. Its structural and legal principles extend to other high-precision texts such as contract clauses, policy reports, clinical documentation, and regulatory filings. By providing a robust framework for evaluating structured texts, our work lays the foundation for establishing reliable evaluation protocols essential for the responsible use of LLMs in high-stakes domains.
\section{Background and Related Work}

\subsection{Specifications in Patent Claims}

\begin{table}[H]
\centering
\begin{tabular}{p{0.94\linewidth}} 
\toprule
\textbf{Claim 1 (Example)} \\
\midrule
1. An extendible trailer tongue comprising: \\
\hspace{1em} a first tubular member; \\
\hspace{1em} a second tubular member telescopically received within the first tubular member; and \\
\hspace{1em} a locking mechanism to secure the second tubular member in a selected position relative to the first tubular member. \\
\bottomrule
\end{tabular}
\caption{A Brief summary example of Claim 1 from the patent titled \textit{``Multiple component headgear system''} \citep{siprut2000headgear}.}
\label{claim1_example}
\end{table}

Patents are specialized legal-technical documents designed to protect intellectual property (IP) rights. Among their components, \textit{claims} play a central role, as they define the precise legal boundaries of protection granted to an invention. In particular, \textit{Claim 1}, which is the first claim, typically serves as the cornerstone of the claim set. It establishes the broadest scope of protection and strongly influences both the validity and enforceability of the patent as a whole \citep{wipo2023manual, merges1997patent}. The breadth of Claim 1 not only constrains subsequent dependent claims but also directly affects the legal and commercial value of the invention.  

Unlike general technical documents, patent claims must strictly adhere to structural, linguistic, and legal requirements. They follow a standardized format mandated by patent offices, employ precise domain-specific terminology, and are drafted with enforceability in mind. Table~\ref{claim1_example} illustrates an example of a well-structured Claim 1, which demonstrates these key requirements:

\begin{itemize}
    \item \textbf{Structural format:} Compliance with strict grammatical and formatting conventions, including numbered clauses, transitional phrases (e.g., ``comprising''), and hierarchical listing of elements. Deviations can render a claim unclear or legally unenforceable.

    \item \textbf{Technical clarity:} Precise definition of technical element and its relationships, ensuring that the scope of the invention is unambiguous to examiners, practitioners, and courts.  
    \item \textbf{Terminology consistency:} Reliance on standardized patent-specific language and consistent referencing of terms, minimizing ambiguity and potential disputes.  
\end{itemize}

Together, these interrelated requirements ensure that Claim 1 functions simultaneously as a detailed technical specification and as a legally binding definition of the scope of the invention. Drafting Claim~1, therefore, requires a delicate and deliberate balance: it must be broad enough to secure meaningful protection and commercial value, while being sufficiently precise and consistent to withstand rigorous legal scrutiny.

%%%%%%%%%%%%%%%%%%%%%%%%%%%%%%%%%%%%%%%%%%%%%%%%%%%%%%%
%%%%%%%%%%%%%%%%%%%%%%%%%%%%%%%%%%%%%%%%%%%%%%%%%%%%%%%
%%%%%%%%%%%%%%%%%%%%%%%%%%%%%%%%%%%%%%%%%%%%%%%%%%%%%%%

\subsection{Patent Generation and Evaluation}

Automating patent generation and evaluation has recently emerged as an important research agenda, yet it remains relatively underexplored. LLMs have been increasingly applied to generate technical and legal content \citep{brown2020language}, showing promise in zero-shot reasoning \citep{kojima2023large} and complex instruction following \citep{Lai2023legal}. However, most existing work has focused on general text generation, without addressing the strict structural and legal constraints unique to patent claims \citep{surden2019artificial}.  

Patent evaluation inherently involves both linguistic and domain-specific dimensions. Traditionally, examiners and attorneys manually review claims, particularly Claim 1, which is labor-intensive and often inconsistent. With the rapid global increase in patent applications, automated evaluation systems must be able to simultaneously assess \textit{technical correctness}, \textit{legal soundness}, and \textit{structural compliance} \citep{surden2019artificial}.  

Nevertheless, evaluating LLM-generated patent claims remains a highly challenging and technically demanding task. Even minor errors in structure, terminology, or legal scope can result in costly litigation or eventual invalidation of the patent. Existing NLG metrics tend to emphasize surface-level linguistic similarity or superficial fluency, but they fail to capture critical patent-specific aspects such as claim scope, legal consistency, structural validity, and the intricate interdependence of claim elements. This limitation strongly underscores the necessity for domain-specific evaluation frameworks that can comprehensively and reliably address the unique requirements of patent claims. 

%%%%%%%%%%%%%%%%%%%%%%%%%%%%%%%%%%%%%%%%%%%%%%%%%%%%%%
%%%%%%%%%%%%%%%%%%%%%%%%%%%%%%%%%%%%%%%%%%%%%%%%%%%%%%
%%%%%%%%%%%%%%%%%%%%%%%%%%%%%%%%%%%%%%%%%%%%%%%%%%%%%%
\subsection{NLG Metrics for Patent Evaluation}
Traditionally, the evaluation of patent texts has relied on metrics using n-grams such as BLEU and ROUGE, which measure lexical overlap between generated and reference texts. However, these approaches overlook semantic relationships and domain-specific nuances, and thus are insufficient for patent claims that demand high levels of legal and technical precision. With the introduction of Transformer-based models, more nuanced evaluation methods became possible. For example, BERTScore leverages contextual embeddings to assess semantic similarity \citep{zhang2019bertscore}, while BLEURT combines BERT’s representational power with task-specific optimization~\citep{sellam2020bleurt}.

More recently, LLM-based evaluation frameworks have emerged, directly employing large language models as evaluators. GPTScore exploits the reasoning capability of LLMs to assess semantic coherence, logical consistency, and contextual appropriateness~\citep{fu-etal-2024-gptscore}. Other studies further highlight the potential of LLM-as-a-Judge~\citep{zheng2023judging}, showing that LLM evaluators can often align more closely with human judgment compared to traditional automatic metrics.

However, these approaches still fail to adequately capture the strict legal and structural constraints inherent in patent claims. Critical aspects such as claim dependency, antecedent basis, and legal clarity remain largely beyond the practical scope of existing evaluation metrics. Unlike general text, patent claims require rigorous structural compliance, precise legal terminology, and systematic domain-specific validation. This persistent gap underscores the urgent need for specialized evaluation frameworks that can comprehensively address both linguistic quality and patent-specific requirements. PatentScore is explicitly designed to fill this void by holistically integrating structural, semantic, and legal dimensions into a unified framework tailored for patent claims.

\begin{figure*}[t]
    \centering
    \includegraphics[width=0.95\textwidth,keepaspectratio]{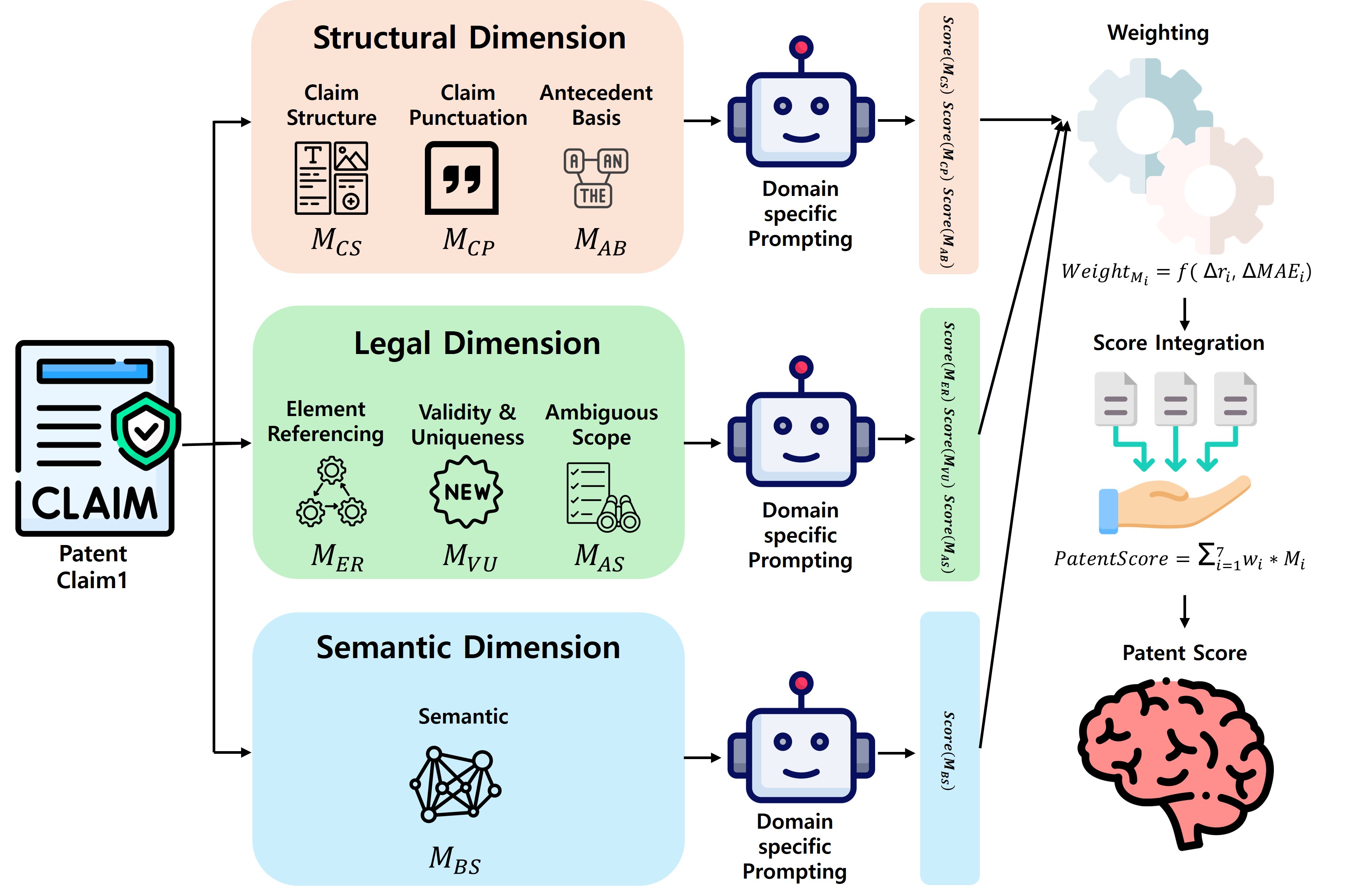}
    \caption{PatentScore Evaluation Framework: decomposition into structural, legal, and semantic dimensions, normalization of metrics, and weighted aggregation for final scoring.}
    \label{fig:patentscore}
\end{figure*}

\section{PatentScore for Patent Claim Evaluation}

\subsection{Evaluation Dimensions of PatentScore}
PatentScore is a multi-dimensional evaluation framework that assesses the quality of patent claims by decomposing them into three independent aspects: structural, legal, and semantic. Each dimension is grounded in domain-specific standards, particularly the WIPO Patent Drafting Manual~\cite{wipo2023manual} and the USPTO~\cite{uspto2023mpep}, and is operationalized through distinct component metrics. These dimensions operate independently, allowing for a detailed diagnosis of various characteristics of a patent claim.

\paragraph{(a) Structural Dimension.}
The \textit{structural dimension} evaluates the syntactic organization and formal composition of patent claims. It focuses on whether the claim conforms to internationally accepted formatting rules that promote legal clarity and technical completeness. This dimension comprises the following three components:
\begin{itemize}
    \item \textbf{Claim Structure} ($M_{CS}$): Checks whether the claim follows the canonical structure, including a preamble, constituent elements, and functional linkages, and logical coherence between interconnected parts.
    \item \textbf{Claim Punctuation} ($M_{CP}$): Verifies proper punctuation use (e.g., commas, semicolons) that affect precise clause separation and legal interpretation.
    \item \textbf{Antecedent Basis} ($M_{AB}$): Assesses the proper usage of definite noun phrases (e.g., ``the module'', ``said component'') that must be supported by clear prior mention.
\end{itemize}

\paragraph{(b) Legal Dimension.}
The \textit{legal dimension} assesses whether the claim satisfies enforceability criteria under patent law, referencing legal doctrine from the USPTO’s MPEP and WIPO standards. Its core metrics include:
\begin{itemize}
    \item \textbf{Element Referencing} ($M_{ER}$): Measures proper dependency and referencing in independent and dependent claims.
    \item \textbf{Validity and Uniqueness} ($M_{VU}$): Evaluates novelty, non-triviality, and absence of internal contradiction.
    \item \textbf{Ambiguous Scope} ($M_{AS}$): Identifies vague or overbroad terms that could undermine legal clarity or enforceability.
\end{itemize}

\paragraph{(c) Semantic Dimension.}
The \textit{semantic dimension} evaluates the extent to which a generated claim preserves the intended meaning and technical context of the original. We adopt BERTScore ($M_{BS}$), which measures semantic similarity using contextual embeddings, due to its robustness in capturing meaning beyond lexical overlap. To ensure compatibility with other dimensions, BERTScore values in $[0, 1]$ are linearly rescaled to a $[1, 5]$ range.

This decomposition enables fine-grained diagnostics of claim quality, enhancing both the reliability and interpretability of LLM-based assessments.

\subsection{Implementation of PatentScore}
To operationalize the three evaluation dimensions, PatentScore employs an LLM-based prompt evaluation scheme. For each metric, a dedicated prompt is designed to guide the model’s attention to the corresponding evaluation criterion. GPT-4o-mini is used as the primary evaluator, and ten independent evaluations are conducted for each metric, with the average score adopted as the final result. This design reduces output variance and ensures stability in scoring.

Different evaluation strategies are applied depending on the nature of each metric. For structural and legal metrics, explicit prompts are used to focus the model on the core evaluation concepts. In contrast, the semantic metric is assessed using BERTScore, which is computed independently. The original $[0, 1]$ range of BERTScore values is linearly rescaled to a $[1, 5]$ range for consistency with other scores. All prompt templates are provided in detail in Appendix~\ref{appendix:prompt}.

The proposed implementation is model-agnostic and can be applied to various LLMs, including GPT-4o-mini and others. Since the evaluation criteria are grounded in internationally recognized patent drafting guidelines (WIPO, USPTO), the framework ensures generalizability and applicability across different contexts.

\subsection{Prompt Design for Claim Evaluation}
In this study, we propose an LLM-adaptable prompt framework for systematic evaluation of patent claims. Our prompt engineering methodology builds upon Chain of Thought~\citep{wei2022chain} reasoning, extending its sequential inference by incorporating prompts that explicitly address structural, legal, and semantic requirements of patent claims. For instance, prompts for structural metrics guide the model to verify claim formatting, while legal prompts focus on compliance with patentability criteria. This expanded prompt structure ensures consistency across evaluations while enabling parallel processing of multiple evaluation dimensions, allowing for a more sophisticated and systematic analysis. Detailed prompts for each evaluation component are provided in Appendix \ref{appendix:prompt}.

\textbf{Component-specific Template Design}
We designed a standardized base template inspired by Chain of Thought, extending its sequential reasoning to a multidimensional framework tailored for patent claim evaluation, enabling both consistency and comprehensive analysis.

\begin{lstlisting}[frame=single]
TASK: \textbf{Patent Claim 1 Assessment}
FOCUS: Component-specific evaluation
OUTPUT FORMAT: Detailed analysis and score (1-5)

\textbf{I. Evaluation Criteria}
Primary Requirements:
- [Component-specific requirement 1]
- [Component-specific requirement 2]
Secondary Requirements:
- [Additional evaluation points]

\textbf{II. Analysis Procedure}
1. Identify relevant claim elements
2. Evaluate against requirements
3. Assign score with justification

\textbf{III. Scoring Rubric}
5: Exceeds all requirements with exceptional clarity
4: Meets all requirements effectively
3: Meets primary requirements adequately
2: Partially meets requirements with deficiencies
1: Fails to meet critical requirements
\end{lstlisting}

\textbf{Implementation by Components}
We customize the templates for structural analysis ($M_{CS}$, $M_{CP}$, $M_{AB}$) and legal compliance ($M_{ER}$, $M_{VU}$, $M_{AS}$) components. For semantic evaluation ($M_{BS}$), we utilize BERTScore, which provides an automated assessment of semantic similarity between generated and reference claims through contextual embeddings. This structured approach, combining specialized prompts for structural and legal assessments with automated semantic evaluation, enables a more reliable and fully reproducible evaluation of patent claim quality across all dimensions of our framework.

\subsection{Score Integration and Weighting}
\label{sec:score_integration}

PatentScore integrates seven metric scores into a final evaluation score by assigning a weight to each component. To determine the weights, we conduct ablation studies that quantify the relative importance of each evaluation metric.

For each component $M_i$, we compute its impact on performance by removing it from the evaluation and observing the changes in Pearson correlation with expert scores ($\Delta r_i = r_{\text{full}} - r_i$) and mean absolute error ($\Delta \text{MAE}_i = \text{MAE}_i - \text{MAE}_{\text{full}}$) relative to the full model. Specifically:

\begin{equation}
\Delta \text{MAE}_i = \text{MAE}_{i} - \text{MAE}_{\text{full}},
\end{equation}
where $\text{MAE}_{\text{full}}$ are the performance metrics using all components, and $\text{MAE}_i$ are the metrics after removing component $M_i$.
The final weight $w_i$ for component $M_i$ is then calculated as:

\begin{equation}
w_i = \frac{\Delta r_i + \Delta \text{MAE}_i}{\sum_{j=1}^{7}(\Delta r_j + \Delta \text{MAE}_j)}.
\end{equation}

To ensure compatibility, each metric score $M_i$ is normalized to the range $[1, 5]$ before aggregation. The final PatentScore is computed as the weighted sum of all component scores:

\begin{equation}
\label{eq:final_patentscore}
\text{PatentScore} = \sum_{i=1}^{7} w_i \cdot M_i.
\end{equation}
This weighted integration reflects the empirically derived contribution of each component and enhances the alignment of PatentScore with expert human evaluations. It rewards dimensions with higher diagnostic power to exert more influence on the final score, while preserving interpretability and fairness across evaluation aspects. 
\section{Data and Verification Measures}
\subsection{Claim Data Generation and Composition}

\begin{table}[h]
\footnotesize
\centering
\renewcommand{\arraystretch}{1.2}
\begin{tabular}{l|p{5cm}}
\hline
\multicolumn{2}{l}{\textbf{Patent Number}: US20170123456\rule{0pt}{2.5ex}} \\
\hline
\textbf{Gold-Claim } & A method for processing biometric data, comprising: receiving sensor data; extracting features; and authenticating a user. \\
\hline
\textbf{LLM-Claim} & A method for biometric processing, comprising: obtaining sensor data; processing features; and performing authentication. \\
\hline
\textbf{Expert Scores} & \begin{tabular}[c]{@{}l@{}}
E1: 3.0 \quad E2: 4.0 \quad E3: 3.0\\
\textit{Mean Score}: 3.33
\end{tabular} \\
\hline
\end{tabular}
\caption{An example of a human-authored patent claim (Gold-Claim) and its GPT-4o-mini-generated (LLM-Claim) counterpart, with expert annotations.}
\label{tab:claimdata}
\end{table}

We use a subset of the HUPD dataset \citep{suzgun2022hupd}, consisting of patents filed in 2016 and 2017, selected for their recency and relevance to modern technological domains. This subset includes the first claim of 400 patents classified under Section A (Human Necessities) and Section G (Physics) of the International Patent Classification (IPC), with 200 claims from each section, ensuring a balanced representation of diverse technical fields. The dataset was curated to support the comparative evaluation of patent claims generated by GPT-4o-mini and those written by human experts. Table~\ref{tab:claimdata} presents an example entry, which serves as the basis for assessing the legal and technical completeness of generated claims, and for validating our proposed sequential evaluation metric. Each entry includes the patent ID, the original human-authored Claim 1, and the corresponding GPT-4o-mini-generated version. The dataset and implementation used in this study are publicly available at \url{https://github.com/Yongmin-Yoo/PatentScore}. 

\subsection{Expert Evaluation}
To verify the reliability of LLM-generated claims and validate the effectiveness of PatentScore, we organized an evaluation panel comprising three domain experts: one legal expert specializing in patent law and two technical experts holding advanced degrees in electrical engineering and computer science, respectively. Each expert independently assessed the generated and reference claims using a 5-point Likert scale (1 = poor, 5 = excellent). The assessment was conducted under blind conditions to prevent bias, and each claim was evaluated on multiple criteria, including \textit{(i)} structural compliance with patent claim conventions (e.g., preamble-body format, proper use of antecedents), \textit{(ii)} clarity and precision in legal phrasing, and \textit{(iii)} completeness and correctness of the technical disclosure.

To ensure reliability and reproducibility, we recorded all individual expert scores (E1–E3) and computed their mean for each claim pair, as summarized in Table~\ref{tab:claimdata}. This multi-expert annotation provides a robust ground truth for benchmarking automatic evaluators and also enables error analysis in cases of inter-rater disagreement. The use of experts from both legal and technical domains ensures that PatentScore is validated against the multi-faceted nature of patent quality, thereby reinforcing its credibility as an evaluation framework.

\subsection{Verification Measures}
To validate the effectiveness of PatentScore, we conducted a comparative evaluation against widely used NLG metrics, including BLEU, ROUGE-L, BERTScore, GPTScore, and Cosine Similarity. As shown in Table \ref{table-metrics}, we assessed alignment with expert judgments using four statistical indicators: Pearson’s $r$, Spearman’s $\rho$, Kendall’s $\tau$, and mean absolute error (MAE). These quantitative measures capture correlation, ranking consistency, and deviation from expert scores, providing a comprehensive basis for evaluating the reliability of PatentScore relative to conventional metrics.

\section{Experiment Results}

\subsection{Claim Generation Quality Analysis}

\begin{table}[h!] 
\centering
\small

\label{tab:metric_comparisons}
\begin{tabular}{lccccc}
\toprule
\textbf{Metric} & \textbf{$r$} & \textbf{$\rho$} & \textbf{$\tau$} & {MAE} \\
\midrule
PatentScore & \textbf{.819} & \textbf{.813} & \textbf{.665} & \textbf{.568} \\
PatentScore Avg. & .818 & .793 & .551 & .568 \\
\midrule
BERTScore & -.161 & -.163 & -.130 & 1.975 \\
GPT Score & .013 & -.004 & -.003 & 1.408 \\
BLEU & -.117 & -.089 & -.065 & 1.744  \\
ROUGE-L & -.159 & -.112 & -.080 & 2.499 \\
Cosine Sim & -.050 & .030 & .024 & 1.946 \\
\bottomrule
\end{tabular}
\caption{The comparison of various metrics using Pearson correlation $r$, Spearman correlation $\rho$, Kendall's Tau $\tau$ and MAE.}
\label{table-metrics}
\end{table}

Our framework demonstrates strong alignment with human expert evaluation, achieving a Pearson correlation coefficient of $r=0.819$ ($p<0.01$) across all metrics, substantially surpassing existing methods. It is noteworthy that the consistent negative correlation shown by traditional metrics such as BERTScore ($r=-0.161$), ROUGE-L ($r=-0.159$), and BLEU ($r=-0.117$), indicating their limitations in patent claim evaluation. The effectiveness of the framework is further validated by its strong performance across multiple correlation measures, including Spearman's rank correlation ($\rho=0.813$) and Kendall's rank correlation ($\tau=0.665$). Beyond correlation coefficients, our approach demonstrates significantly lower error rates ($MAE=0.568$) compared to existing metrics, which exhibit substantially higher error values (e.g., $MAE=2.499$ for ROUGE-L).

Our results reveal that commonly used metrics show limited effectiveness in capturing the nuances of patent claim quality, as indicated by their negative correlations with expert judgments. While metrics like BERTScore and ROUGE-L show negative or near-zero correlation with expert scores, PatentScore exhibits strong positive correlation, indicating its suitability for the legal and structural nuances of patent claims. The consistently superior performance of our framework across diverse criteria establishes it as a robust and reliable method for patent claim assessment. To assess the generalizability of PatentScore beyond a single generation model, we conducted additional evaluations using GPT-generated claims from two distinct LLMs, Claude-3.5-Haiku and Gemini-1.5-Flash on the same 400-pair dataset under identical evaluation conditions. PatentScore achieved Pearson correlation coefficients of 0.745 and 0.731 on these models, respectively, indicating consistently strong alignment with expert annotations across diverse LLMs. These results reinforce the robustness and cross-model applicability of PatentScore as a reliable evaluator for patent claim quality.

\subsection{Quality Comparison of Expert Evaluation on Generated Claims}
The Pearson correlation coefficients between the ratings by three experts range from 0.795 to 0.847, demonstrating their high correlation and indicating that the evaluators followed similar standards in assessing claim quality. Cronbach's $\alpha$ is 0.931, reflecting a very high level of internal consistency in their judgments across different claims. Furthermore, the ICC(3,k) value of 0.928 confirms excellent inter-rater agreement, while Krippendorff's $\alpha$ of 0.784 suggests a substantial level of reliability even under more conservative assumptions. Taken together, these findings statistically validate that the proposed evaluation framework captures rigorous and consistent criteria, ensuring high levels of reliability and reproducibility in expert-based assessments. Importantly, such strong agreement among evaluators provides a solid foundation for benchmarking automated metrics, as it establishes that human judgments themselves are both stable and trustworthy, thereby reinforcing the credibility of subsequent comparisons between expert ratings and model-generated scores.

\subsection{Ablation Study} 
\label{sec:ablation}

\renewcommand{\arraystretch}{1.2} 
\begin{table}[H] 
\centering
\resizebox{0.98\columnwidth}{!}{%
\begin{tabular}{lccccc}
\hline
\textbf{Model\rule{0pt}{2.5ex}} & \textbf{Corr.} ($r$) & \textbf{MAE} & \textbf{Perf. Drop} (\%) & \textbf{Weight} \\
\hline
Baseline ($\ddagger$) & 0.818 & 0.568 & - & 1.0 \\
\textbf{- $M_{CS}$} & 0.735 & 0.675 & 10.1 (18.8) & 0.166  \\
\textbf{- $M_{CP}$}& 0.723 & 0.667 & 11.6 (17.4) & 0.171  \\
\textbf{- $M_{AB}$} & 0.731 & 0.671 & 10.6 (18.1) & 0.167 \\
\textbf{- $M_{ER}$} & 0.734 & 0.670 & 10.3 (18.0) & 0.163\\
\textbf{- $M_{VU}$} & 0.742 & 0.674 & 9.3 (18.7) & 0.159  \\
\textbf{- $M_{AS}$} & 0.729 & 0.677 & 10.9 (19.2) & 0.173  \\
\textbf{- $M_{BS}$} & 0.817 & 0.569 & 0.1 (0.2) & 0.001 \\
\hline
\end{tabular}%
}
\footnotesize
\textbf{Notes:} Numbers in parentheses indicate MAE increases. The baseline model ($\ddagger$) uses equal weights for all components.
\caption{The comparison of metric contributions via ablation analysis.}
\vspace{1mm}
\label{tab:ablation-performance}
\end{table}

We conducted an ablation study to analyze the impact of each metric on the overall framework performance, as shown in Table~\ref{tab:ablation-performance}. The baseline model, which assigns equal weights (i.e., $w_i = 1/7$) to all seven components, achieves a correlation of 0.818 with MAE of 0.568. Further, component-wise analysis reveals:

\begin{itemize}
  \item \textbf{\textit{Structural metrics}}: The removal of structural components, including claim structure ($M_{CS}$), punctuation pattern ($M_{CP}$), and antecedent basis ($M_{AB}$), led to correlation drops of 10.1\%, 11.6\%, and 10.6\%, respectively. These substantial decreases highlight the central role of structural compliance in patent claim evaluation. In particular, the antecedent basis metric, which ensures logical referential consistency across claim elements, emerges as a key determinant of expert-perceived quality.  

  \item \textbf{\textit{Legal metrics}}: Excluding legal criteria, element referencing ($M_{ER}$), validity and uniqueness ($M_{VU}$), and ambiguous scope ($M_{AS}$), resulted in performance reductions of 10.3\%, 9.3\%, and 10.9\%. These findings indicate that legal soundness is not a marginal factor but a core dimension in assessing claim quality. The ambiguous scope metric shows that evaluators strongly penalize imprecise or overly broad language, reaffirming the necessity of explicit legal rigor in automated evaluation.  

  \item \textbf{\textit{Semantic metric}}: In contrast, the exclusion of the semantic metric BERTScore ($M_{BS}$) produced only a negligible decline of 0.1\%. This suggests that while semantic similarity is not independently decisive, it provides complementary value by reinforcing lexical and terminological consistency, particularly in contexts involving emerging or domain-specific technical vocabulary.  

\end{itemize}

Following the final weights in Eqn.~\ref{eqn_weight}, we can calculate the patent score, where each weight $W$ reflects the relative importance of corresponding metrics derived from the ablation study.

\begin{equation}
\begin{split}
\label{eqn_weight}
\textit{PatentScore} = & \, 0.166 \cdot M_{CS} + 0.171 \cdot M_{CP} + \\
                      & \,  0.163 \cdot M_{ER} + 0.167 \cdot M_{AB} + \\
                      & \,  0.159 \cdot M_{VU} + 0.173 \cdot M_{AS} + \\
                      & \,  0.001 \cdot M_{BS}
\end{split}
\end{equation}

These findings highlight the minimal impact of BERTScore on overall framework performance. Removing BERTScore results in only a 0.1\% decrease, indicating that semantic metrics play a complementary rather than critical role in patent claim evaluation.  
In contrast, structural and legal metrics demonstrate substantial influence, with performance drops of 9.3\% to 11.6\% when excluded underscoring the importance of structural validity and legal compliance in determining claim quality.  
While modern LLMs exhibit strong semantic consistency and accuracy, they still require explicit guidance to meet structural and legal requirements. This balance of semantic strength and structural/legal support affirms the necessity of a comprehensive evaluation framework like PatentScore.

\subsection{Key Findings}
\textbf{\textit{Limitations of Existing Metrics}}:  
Conventional text evaluation metrics (BLEU, ROUGE-L, and BERTScore) exhibit negative correlations ($r = -0.117 \sim -0.161$) when applied to patent claim evaluation, underscoring their inadequacy for this specialized domain. BLEU's reliance on n-gram overlap fails to capture logical dependencies, and ROUGE-L prioritizes surface-level similarity at the expense of legal and technical coherence. These limitations clearly highlight the need for a domain-tailored evaluation framework.  

\textbf{\textit{Balanced Component Evaluation}}:  
Our ablation study reveals comparable weights for structural ($0.166-0.171$) and legal components ($0.159-0.173$), demonstrating that both dimensions contribute equally to robust patent claim assessment. The 10.6\% performance drop from omitting the antecedent basis metric ($M_{AB}$) illustrates how critical cross-referential clarity is in expert judgments. This finding confirms that a multi-faceted evaluation approach, rather than reliance on a single dimension, is essential for capturing claim quality.  

\textbf{\textit{Semantic Analysis Role}}:  
Although BERTScore contributes only marginal independent impact (weight: $0.001$), its inclusion plays a complementary role by reinforcing domain-specific terminology consistency, particularly for emerging technology terms. This demonstrates that semantic alignment alone is insufficient but still valuable as a supporting layer. Together with structural and legal measures, it enhances the comprehensiveness of PatentScore, ensuring that both form and meaning are adequately captured.  

\section{Conclusion}

This research proposed PatentScore, a multi-dimensional framework that systematically integrates structural, legal, and semantic metrics for evaluating patent claims. Unlike conventional NLG metrics that fail to capture domain-specific requirements, PatentScore explicitly incorporates legal soundness and structural validity, achieving strong alignment with expert judgments ($r = 0.819$) and significantly reducing error rates.

By bridging legal expertise with language modeling, PatentScore provides a reliable, transparent, and practically useful evaluation foundation, establishing a pathway for more trustworthy and innovative applications of AI in the patent domain.
The experimental results show that structural and legal factors play a decisive role in assessing claim quality, while semantic metrics provide complementary value by reinforcing terminology consistency. Through expert-based validation, comparative benchmarking, and ablation studies, PatentScore is demonstrated to be a rigorous, interpretable, and reliable tool for evaluating LLM-generated claims.

While the proposed framework is developed with a focus on patent claims, it can be easily extended to other high-precision document types that require structural constraints and legal clarity, such as contract clauses, policy reports, and regulatory compliance in clinical documentation. This contributes to the broader scope of establishing domain-specific evaluation criteria for the trustworthy deployment of LLMs in high-stakes settings.

\section*{Limitations}
\label{appendix:limitation}

This study has several limitations, which also suggest promising directions for future research.

First, the dataset used in this study is restricted to a specific technical domain and filing period. Expanding the scope to include a broader range of technologies and time frames would enhance the generalizability of the framework.

Second, PatentScore primarily evaluates intrinsic aspects of claim quality, such as structural, legal, and semantic dimensions, but does not address extrinsic properties like novelty or distinctiveness. Future work could incorporate comparative modules that measure overlap with prior art, enabling assessments of uniqueness and scope validity, which are critical in real-world patent examination.

Third, while PatentScore is designed for evaluating LLM-generated claims, its modular architecture allows adaptation to human-written patents as well. This extension would enable automated analysis and feedback on structural coherence, clarity, and legal soundness in expert-authored claims.

Fourth, integrating legal knowledge graphs into the generation and evaluation pipeline could further improve factual accuracy and contextual alignment. Such integration would strengthen the framework’s ability to handle complex dependencies across legal and technical contexts.

\section*{Ethical Considerations} 
In conducting this research, we ensured the highest ethical standards were upheld. All data used in this study, including GPT-4o-mini generated claims and the HUPD dataset, were obtained from publicly available sources and used in compliance with the respective licenses, such as the CC BY License. Human evaluators who participated in the evaluation process provided voluntary and informed consent before contributing to the study. Their identities remain anonymous to protect their privacy and confidentiality.

The study does not involve any personally identifiable information or sensitive data, and it strictly adheres to ethical guidelines for research involving human evaluators. Furthermore, we acknowledge the potential for misuse of AI in automating patent generation and emphasize the importance of developing AI tools that align with legal and ethical standards. The proposed framework, PatentScore, aims to improve transparency, reliability, and accountability in AI-generated patent claims.

\section*{Acknowledgments} 
This work was supported by the Australian Research Council (ARC) under the following projects: 
LP230201022 (Ethical Enterprise Representations for Personalised Sustainable Finance, Australian Research Council Linkage Grants), 
DP240102050 (Data Complexity and Uncertainty-Resilient Deep Variational Learning, Australian Research Council Discovery Grants), 
and LE240100131 (Federated Omniverse Facilities for Smart Digital Futures, Australian Research Council Linkage Infrastructure, Equipment and Facilities Grants). 

We thank the creators of the HUPD dataset (CC BY License) and OpenAI for access to GPT, both of which were essential resources for this study. 
We are also grateful to the three evaluators for their valuable contributions, careful assessments, and voluntary participation after providing informed consent. 
Their expertise and feedback were instrumental in validating the evaluation framework. 
Finally, we acknowledge the broader research community for developing the open-source tools and prior works that served as the foundation for this study.

\bibliography{anthology,custom}
\bibliographystyle{acl_natbib}

\appendix
\appendix

\section{Appendix}
\label{appendix:prompt}

%\label{sec:appendix}

\subsection{Prompts for Experiments}
%\label{appendix:prompt}
\subsubsection{Prompts for Claim Structure Evaluation}

\subsection*{Name: Patent Claim Structure}

\subsubsection*{I. Overview}
\textbf{Definition:} A patent claim's composition consisting of three essential parts (preamble, transitional phrase, and elements), which verify whether all these elements are properly positioned.  

\textbf{Evaluation Purpose:} To systematically assess the structural completeness and proper organization of a patent claim to ensure clear and enforceable rights.

\subsubsection*{II. Key Evaluation Points}
\begin{enumerate}
    \item Preamble identification and content validation
    \item Transitional phrase appropriateness and placement
\end{enumerate}

\subsubsection*{III. Important Notes}
\begin{itemize}
    \item Ensure all three essential parts exist.
    \item Check proper element positioning.
\end{itemize}

\subsubsection*{IV. Standard Examples}
\begin{quote}
"A device for processing data, comprising:  
a memory configured to store data; 
a processor coupled to the memory; and  
an interface connected to the processor."
\end{quote}

\subsubsection*{V. Evaluation Procedure}
\begin{enumerate}
    \item \textbf{Content Identification}  \\
    \textbf{Input:} Claim 1 requiring evaluation  \\
    \textbf{Output:} Checklist, mapping, and element classification \\ 
    \textbf{Q:} Have all structural elements been identified? \\ 
    \textbf{Tasks:}
    \begin{itemize}
        \item Carefully read the provided Claim 1 to identify:
        \begin{itemize}
            \item Preamble
            \item Transitional phrase
            \item Elements/components
        \end{itemize}
        \item Verify presence of all essential parts:
        \begin{itemize}
            \item Preamble completeness
            \item Appropriate transitional phrase
            \item All necessary elements
        \end{itemize}
        \item Create mapping of identified elements.
    \end{itemize}

    \item \textbf{Critical Element Verification}  \\
    \textbf{Input:} Identified elements and relationships  \\
    \textbf{Output:} Verification report of required elements \\ 
    \textbf{Q:} Do the essential elements meet requirements?\\  
    \textbf{Tasks:}
    \begin{itemize}
        \item Confirm essential structural elements:
        \begin{itemize}
            \item Clear and complete preamble statement
            \item Standard transitional phrase usage
            \item Properly structured body elements
        \end{itemize}
        \item Verify element relationships:
        \begin{itemize}
            \item Proper hierarchical structure
        \end{itemize}
        \item Remember missing or inappropriate items
    \end{itemize}

    \item \textbf{Format Compliance Assessment}  \\
    \textbf{Input:} Element verification results  \\
    \textbf{Output:} Compliance evaluation report \\ 
    \textbf{Q:} Does it comply with format requirements? \\ 
    \textbf{Tasks:}
    \begin{itemize}
        \item Check format compliance:
        \begin{itemize}
            \item Indentation and spacing
            \item Punctuation usage (colons, semicolons)
            \item Conjunction placement
        \end{itemize}
        \item Identify rule violations
        \item Remember formatting errors
    \end{itemize}

    \item \textbf{Standard-Based Comparative Analysis}  \\
    \textbf{Input:} Compliance assessment results \\ 
    \textbf{Output:} Comparative analysis report \\ 
    \textbf{Q:} How well does it match standards/examples? \\ 
    \textbf{Tasks:}
    \begin{itemize}
        \item Compare with standard claim structures:
        \begin{itemize}
            \item Similar technology field examples
            \item Accepted structural patterns
        \end{itemize}
        \item Identify structural deviations
        \item Remember errors
    \end{itemize}

    \item \textbf{Final Scoring}  \\
    \textbf{Input:} Analysis results from previous steps \\ 
    \textbf{Output:} Final numerical score \\ 

    \textbf{Patent Claim Structure Scoring (1-5 points):}
    \begin{itemize}
        \item 1 point (\textless 20\%):  
        \begin{quote}
        Example: "Processing data with memory and processor." \\ 
        \textbf{Issues:} No clear preamble, no transitional phrase, unstructured elements, missing punctuation, no formatting.
        \end{quote}

        \item 2 points (20-40\%):  
        \begin{quote}
        Example: "A data processing device having memory for storing data processor for processing interface for connecting." \\ 
        \textbf{Issues:} Weak preamble, missing transitional phrase, poor organization, incorrect punctuation, basic elements only.
        \end{quote}

        \item 3 points (41-60\%):  
        \begin{quote}
        Example: "A device for processing data comprising: a memory that stores data a processor connected to memory an interface; wherein the processor processes data." \\ 
        \textbf{Issues:} Basic preamble, has transitional phrase, basic formatting, some punctuation errors, basic relationships.
        \end{quote}

        \item 4 points (61-80\%):  
        \begin{quote}
        Example: "A device for processing data, comprising: a memory configured to store data; a processor coupled to the memory; an interface connected to the processor; wherein the interface transmits data." \\ 
        \textbf{Strengths:} Clear preamble, correct transitional phrase, good formatting, minor structural issues, clear relationships.
        \end{quote}

        \item 5 points (\textgreater 80\%):  
        \begin{quote}
        Example: "A device for processing data, comprising: a memory configured to store input data; a processor coupled to the memory and configured to process data; an interface coupled to the processor; wherein the processor updates rules based on feedback." \\ 
        \textbf{Strengths:} Perfect structure, proper formatting, clear hierarchy, complete elements, professional composition.
        \end{quote}
    \end{itemize}
\end{enumerate}

\subsubsection{Prompts for Punctuation in Patent Claims}

\subsection*{Name: Punctuation in Patent Claim}

\subsubsection*{I. Overview}
\textbf{Definition:} Evaluation of proper use and placement of essential punctuation marks in patent claims (colon, semicolons, "and", period).  

\textbf{Evaluation Purpose:} To systematically assess the accuracy of punctuation usage to ensure clarity and readability of the claim.

\subsubsection*{II. Key Evaluation Points}
\begin{enumerate}
    \item Essential punctuation mark validation
    \item Appropriate punctuation placement
\end{enumerate}

\subsubsection*{III. Important Notes}
\begin{itemize}
    \item Check for colon (:) after transitional phrase.
    \item Check for semicolons (;) after elements.
    \item Check for "and" before the last element.
    \item Check for period (.) at the end of claim.
\end{itemize}

\subsubsection*{IV. Standard Examples}
\begin{quote}
"A device for processing data, comprising:  
a memory configured to store data;  
a processor coupled to the memory; and  
an interface connected to the processor."
\end{quote}

\subsubsection*{V. Evaluation Procedure}
\begin{enumerate}
    \item \textbf{Content Identification}  \\
    \textbf{Input:} Claim 1 requiring evaluation \\ 
    \textbf{Output:} Checklist, mapping, and punctuation classification \\ 
    \textbf{Q:} Have all punctuation marks been identified? \\ 
    \textbf{Tasks:}
    \begin{itemize}
        \item Carefully read the provided Claim 1 to identify:
        \begin{itemize}
            \item Colon location
            \item Semicolon locations
            \item "And" location
            \item Period location
        \end{itemize}
        \item Verify presence of all essential punctuation:
        \begin{itemize}
            \item Colon after transition
            \item Semicolons after each element
            \item "And" before last element
            \item Period at claim end
        \end{itemize}
        \item Create mapping of identified punctuation
    \end{itemize}

    \item \textbf{Critical Element Verification}  \\
    \textbf{Input:} Identified punctuation and locations \\ 
    \textbf{Output:} Essential punctuation verification report \\ 
    \textbf{Q:} Do the essential punctuation marks meet requirements? \\ 
    \textbf{Tasks:}
    \begin{itemize}
        \item Confirm essential punctuation:
        \begin{itemize}
            \item Appropriate colon usage
            \item Appropriate semicolon usage
            \item Appropriate "and" usage
        \end{itemize}
        \item Verify punctuation locations:
        \begin{itemize}
            \item Exact position of each mark
        \end{itemize}
        \item Remember missing or inappropriate items
    \end{itemize}

    \item \textbf{Format Compliance Assessment}  \\
    \textbf{Input:} Punctuation verification results \\ 
    \textbf{Output:} Compliance evaluation report \\ 
    \textbf{Q:} Does punctuation comply with format requirements? \\ 
    \textbf{Tasks:}
    \begin{itemize}
        \item Check punctuation format compliance:
        \begin{itemize}
            \item Spacing around punctuation
            \item Consistency in usage
        \end{itemize}
        \item Remember formatting errors
    \end{itemize}

    \item \textbf{Standard-Based Comparative Analysis} \\ 
    \textbf{Input:} Compliance assessment results \\ 
    \textbf{Output:} Comparative analysis report \\ 
    \textbf{Q:} How well does it match standards/examples? \\ 
    \textbf{Tasks:}
    \begin{itemize}
        \item Compare with standard punctuation usage:
        \begin{itemize}
            \item Standard case patterns
            \item Accepted usage practices
        \end{itemize}
        \item Identify punctuation deviations
        \item Remember errors
    \end{itemize}

    \item \textbf{Final Scoring}  \\
    \textbf{Input:} Analysis results from previous steps \\ 
    \textbf{Output:} Final numerical score \\ 
    \textbf{Patent Claim Punctuation Scoring (1-5 points):}
    \begin{itemize}
        \item 1 point (\textless 20\%):  
        \begin{quote}
        Example: "A device with memory processor and interface."\\  
        \textbf{Issues:} No colon after transition, no semicolons between elements, missing "and", missing period, no spacing after marks.
        \end{quote}

        \item 2 points (20-40\%):  
        \begin{quote}
        Example: "A device comprising memory; processor; interface."\\  
        \textbf{Issues:} Missing colon after transition, inconsistent semicolon use, missing "and", has period, poor spacing.
        \end{quote}

        \item 3 points (41-60\%):  
        \begin{quote}
        Example: "A device comprising: memory; processor; and interface." \\ 
        \textbf{Issues:} Has colon, some semicolons, has "and", missing period, basic spacing.
        \end{quote}

        \item 4 points (61-80\%):  
        \begin{quote}
        Example: "A device, comprising: memory; processor; and interface." \\ 
        \textbf{Strengths:} Proper colon, most semicolons correct, proper "and", has period, minor spacing issues.
        \end{quote}

        \item 5 points (\textgreater 80\%):  
        \begin{quote}
        Example: "A device for processing data, comprising: a memory configured to store data; a processor coupled to the memory; and an interface connected to the processor." \\ 
        \textbf{Strengths:} Perfect colon placement, all semicolons correct, proper "and" placement, correct period, perfect spacing.
        \end{quote}
    \end{itemize}
\end{enumerate}

\subsubsection{Claim Inconsistent Element Referencing}

\subsection*{Name: Claim Inconsistent Element Referencing}

\subsubsection*{I. Overview}
\textbf{Definition:} Assessment of consistency in referring to claim components (e.g., touchscreen display, processor, battery), ensuring each element maintains uniform terminology throughout the claim and avoids ambiguous references.  

\textbf{Evaluation Purpose:} To ensure each component in the claim is consistently referenced using precise terminology, maintaining clarity and avoiding confusion in element relationships.

\subsubsection*{II. Key Evaluation Points}
\begin{enumerate}
    \item Terminology consistency verification
    \item Reference clarity assessment
\end{enumerate}

\subsubsection*{III. Important Notes}
\begin{itemize}
    \item Check for consistent use of component terms
    \item Verify clear reference relationships
    \item Confirm no ambiguous references
\end{itemize}

\subsubsection*{IV. Standard Examples}
\begin{quote}
"A device, comprising:  
a touchscreen display;  
a processor connected to the touchscreen display; and  
a battery powering the processor and the touchscreen display."
\end{quote}

\subsubsection*{V. Evaluation Procedure}
\begin{enumerate}
    \item \textbf{Content Identification}  \\
    \textbf{Input:} Claim 1 requiring evaluation \\ 
    \textbf{Output:} Checklist, mapping, and reference classification \\ 
    \textbf{Q:} Have all element references been identified? \\ 
    \textbf{Tasks:}
    \begin{itemize}
        \item Carefully read the provided Claim 1 to identify:
        \begin{itemize}
            \item Each component term
            \item Each reference to components
            \item Reference relationships
        \end{itemize}
        \item Verify component terminology:
        \begin{itemize}
            \item Initial component definitions
            \item Subsequent references
            \item Reference consistency
        \end{itemize}
        \item Create mapping of identified references
    \end{itemize}

    \item \textbf{Critical Element Verification}  \\
    \textbf{Input:} Identified references and relationships \\ 
    \textbf{Output:} Reference verification report \\ 
    \textbf{Q:} Do the reference terms meet requirements? \\ 
    \textbf{Tasks:}
    \begin{itemize}
        \item Confirm reference consistency:
        \begin{itemize}
            \item Uniform terminology use
            \item Clear reference connections
            \item Proper reference terms
        \end{itemize}
        \item Verify reference accuracy:
        \begin{itemize}
            \item Correct component references
        \end{itemize}
        \item Remember inconsistent or improper items
    \end{itemize}

    \item \textbf{Format Compliance Assessment}  \\
    \textbf{Input:} Reference verification results \\ 
    \textbf{Output:} Compliance evaluation report \\ 
    \textbf{Q:} Does reference usage comply with format requirements? \\ 
    \textbf{Tasks:}
    \begin{itemize}
        \item Check reference format compliance:
        \begin{itemize}
            \item Reference term consistency
            \item Reference terminology clarity
        \end{itemize}
        \item Remember formatting errors
    \end{itemize}

    \item \textbf{Standard-Based Comparative Analysis} \\ 
    \textbf{Input:} Compliance assessment results \\ 
    \textbf{Output:} Comparative analysis report \\ 
    \textbf{Q:} How well does it match standards/examples? \\ 
    \textbf{Tasks:}
    \begin{itemize}
        \item Compare with standard reference patterns:
        \begin{itemize}
            \item Standard reference cases
            \item Accepted reference practices
        \end{itemize}
        \item Identify reference deviations
        \item Remember errors
    \end{itemize}

    \item \textbf{Final Scoring}  \\
    \textbf{Input:} Analysis results from previous steps \\ 
    \textbf{Output:} Final numerical score \\ 
    \textbf{Patent Claim Element Reference Consistency Scoring (1-5 points):}
    \begin{itemize}
        \item 1 point (\textless 20\%):  
        \begin{quote}
        Example: "A device including a screen connected to CPU, wherein the display is connected to processor, and a battery powering the touchscreen." \ 
        \textbf{Issues:} Different terms for same element (screen/display/touchscreen), incorrect reference terms (CPU/processor), unclear reference relationships, serious inconsistencies, confusing element relationships.
        \end{quote}

        \item 2 points (20-40\%):  
        \begin{quote}
        Example: "A touchscreen display; a processor connected to the screen; and a battery powering the CPU and display." \\ 
        \textbf{Issues:} Some inconsistent terminology, mixed reference terms, unclear reference relationships, major inconsistencies, basic element mentions only.
        \end{quote}

        \item 3 points (41-60\%):  
        \begin{quote}
        Example: "A touchscreen display; a processor connected to the touchscreen display; and a battery powering the processor and display." \ 
        \textbf{Issues:} Basic consistency, some missing references, partially clear relationships, some inconsistencies, basic reference structure.
        \end{quote}

        \item 4 points (61-80\%):  
        \begin{quote}
        Example: "A touchscreen display; a processor connected to said touchscreen display; and a battery powering said processor and said touchscreen display." \\ 
        \textbf{Strengths:} Mostly consistent terminology, proper reference terms, clear reference relationships, minor inconsistencies, clear element relationships.
        \end{quote}

        \item 5 points (\textgreater 80\%):  
        \begin{quote}
        Example: "A device, comprising: a touchscreen display; a processor operatively connected to said touchscreen display; and a battery configured to power said processor and said touchscreen display." \\ 
        \textbf{Strengths:} Perfect terminology consistency, accurate reference terms, clear reference relationships, professional composition, perfect element relationships.
        \end{quote}
    \end{itemize}
\end{enumerate}

\subsubsection{Patent Claim Antecedent Basis}

\subsection*{Name: Patent Claim Antecedent Basis}

\subsubsection*{I. Overview}
\textbf{Definition:} Assessment of proper introduction and subsequent referencing of claim elements, ensuring each "the" reference has a corresponding previous introduction with "a" or "an".  

\textbf{Evaluation Purpose:} To ensure clear and legally proper element referencing by confirming proper establishment of antecedent basis for all claim elements.

\subsubsection*{II. Key Evaluation Points}
\begin{enumerate}
    \item Article usage verification (\textit{a}, \textit{an}, \textit{the})
    \item First mention validation
\end{enumerate}

\subsubsection*{III. Important Notes}
\begin{itemize}
    \item Check first introduction uses "a" or "an"
    \item Verify subsequent references use "the"
    \item Confirm all "the" references have prior introduction
\end{itemize}

\subsubsection*{IV. Standard Examples}
\begin{quote}
"A device, comprising:  
a display;  
a processor connected to the display; and  
a battery, wherein the battery powers the processor and the display."
\end{quote}

\subsubsection*{V. Evaluation Procedure}
\begin{enumerate}
    \item \textbf{Content Identification}  \\
    \textbf{Input:} Claim 1 requiring evaluation \\ 
    \textbf{Output:} Checklist, mapping, and article usage classification \\ 
    \textbf{Q:} Have all article usages been identified? \\ 
    \textbf{Tasks:}
    \begin{itemize}
        \item Carefully read the provided Claim 1 to identify:
        \begin{itemize}
            \item First mentions (\textit{a}, \textit{an})
            \item Subsequent references (\textit{the})
            \item Reference relationships
        \end{itemize}
        \item Verify article usage:
        \begin{itemize}
            \item Initial introductions
            \item Subsequent references
            \item Reference consistency
        \end{itemize}
        \item Create mapping of identified articles.
    \end{itemize}

    \item \textbf{Critical Element Verification} \\ 
    \textbf{Input:} Identified articles and relationships \\ 
    \textbf{Output:} Article usage verification report \\ 
    \textbf{Q:} Do the article usages meet requirements?\\  
    \textbf{Tasks:}
    \begin{itemize}
        \item Confirm proper introduction:
        \begin{itemize}
            \item Proper first mention articles
            \item Proper subsequent articles
        \end{itemize}
        \item Verify reference accuracy:
        \begin{itemize}
            \item Match between references
        \end{itemize}
        \item Remember improper article usage
    \end{itemize}

    \item \textbf{Format Compliance Assessment}  \\
    \textbf{Input:} Article verification results \\ 
    \textbf{Output:} Compliance evaluation report \\ 
    \textbf{Q:} Does article usage comply with requirements? \\ 
    \textbf{Tasks:}
    \begin{itemize}
        \item Check article usage compliance:
        \begin{itemize}
            \item Article sequence
            \item Reference clarity
        \end{itemize}
        \item Remember formatting errors
    \end{itemize}

    \item \textbf{Standard-Based Comparative Analysis} \\ 
    \textbf{Input:} Compliance assessment results \\ 
    \textbf{Output:} Comparative analysis report \\ 
    \textbf{Q:} How well does it match standards/examples? \\ 
    \textbf{Tasks:}
    \begin{itemize}
        \item Compare with standard article usage:
        \begin{itemize}
            \item Standard reference patterns
            \item Accepted article practices
        \end{itemize}
        \item Identify article usage deviations
        \item Remember errors
    \end{itemize}

    \item \textbf{Final Scoring}  \\
    \textbf{Input:} Analysis results from previous steps \\ 
    \textbf{Output:} Final numerical score \\ 
    \textbf{Patent Claim Antecedent Basis Scoring (1-5 points):}
    \begin{itemize}
        \item 1 point (\textless 20\%):  
        \begin{quote}
        Example: "The display and the CPU, wherein the memory connects to the processor." \\ 
        \textbf{Issues:} Improper first mentions, references without antecedent basis, incorrect article usage, severe antecedent basis issues, confusing relationships.
        \end{quote}

        \item 2 points (20-40\%):  
        \begin{quote}
        Example: "The display and a processor, wherein the CPU connects to memory."  \\
        \textbf{Issues:} Some improper first mentions, incomplete antecedent basis, incorrect article usage, major antecedent basis issues, basic elements only.
        \end{quote}

        \item 3 points (41-60\%):  
        \begin{quote}
        Example: "A display; the processor connected to the display; and the memory connected to the processor." \\ 
        \textbf{Issues:} Basic antecedent basis, some incorrect articles, some unclear references, some antecedent basis issues, basic reference structure.
        \end{quote}

        \item 4 points (61-80\%):  
        \begin{quote}
        Example: "A display; a processor connected to the display; and the processor connected to a memory." \\ 
        \textbf{Strengths:} Mostly proper antecedent basis, proper article usage, clear references, minor antecedent basis issues, clear element relationships.
        \end{quote}

        \item 5 points (\textgreater 80\%):  
        \begin{quote}
        Example: "A display; a processor connected to the display; a memory; and a battery connected to the processor and the memory." \\ 
        \textbf{Strengths:} Perfect antecedent basis, accurate article usage, clear references, professional composition, perfect element relationships.
        \end{quote}
    \end{itemize}
\end{enumerate}

\subsubsection{Claim Validity \& Uniqueness}

\subsection*{Name: Claim Validity \& Uniqueness}

\subsubsection*{I. Overview}
\textbf{Definition:} Evaluation of whether each claim component contributes new technical features and specific elements to the invention, without contradictions or unnecessary repetitions of earlier components.  

\textbf{Evaluation Purpose:} To ensure the claim maintains its validity and uniqueness by differentiating each part and upholding legal strength within the patent framework.

\subsubsection*{II. Key Evaluation Points}
\begin{enumerate}
    \item Technical feature uniqueness
    \item Component differentiation
\end{enumerate}

\subsubsection*{III. Important Notes}
\begin{itemize}
    \item Verify each component adds unique value
    \item Check for technical contradictions
    \item Confirm no unnecessary repetitions
\end{itemize}

\subsubsection*{IV. Standard Examples}
\begin{quote}
"A data processing device, comprising:  
a unique processing unit having specific computational capabilities;  
a specialized memory unit configured to enhance processing speed; and  
a novel interface designed to optimize data transfer."
\end{quote}

\subsubsection*{V. Evaluation Procedure}
\begin{enumerate}
    \item \textbf{Content Identification}  \\
    \textbf{Input:} Claim 1 requiring evaluation \\ 
    \textbf{Output:} Checklist, mapping, and uniqueness classification \\ 
    \textbf{Q:} Have all unique elements been identified?\\  
    \textbf{Tasks:}
    \begin{itemize}
        \item Carefully read the provided Claim 1 to identify:
        \begin{itemize}
            \item Technical features
            \item Unique elements
            \item Specific contributions
        \end{itemize}
        \item Verify element uniqueness:
        \begin{itemize}
            \item Technical differentiation
            \item Specific features
        \end{itemize}
        \item Create mapping of unique features
    \end{itemize}

    \item \textbf{Critical Element Verification}  \\
    \textbf{Input:} Identified elements and features \\ 
    \textbf{Output:} Validity verification report  \\
    \textbf{Q:} Do the elements contribute unique value?  \\
    \textbf{Tasks:}
    \begin{itemize}
        \item Confirm technical contribution:
        \begin{itemize}
            \item Novel features
            \item Specific improvements
        \end{itemize}
        \item Verify element differentiation:
        \begin{itemize}
            \item Technical distinctions
        \end{itemize}
        \item Remember contradictions or repetitions
    \end{itemize}

    \item \textbf{Format Compliance Assessment}  \\
    \textbf{Input:} Verification results \\ 
    \textbf{Output:} Compliance evaluation report \\ 
    \textbf{Q:} Does the claim maintain validity requirements? \\ 
    \textbf{Tasks:}
    \begin{itemize}
        \item Check validity compliance:
        \begin{itemize}
            \item Technical consistency
            \item Feature uniqueness
        \end{itemize}
        \item Remember validity issues
    \end{itemize}

    \item \textbf{Standard-Based Comparative Analysis} \\ 
    \textbf{Input:} Compliance assessment results  \\
    \textbf{Output:} Comparative analysis report \\ 
    \textbf{Q:} How well does it establish uniqueness? \\ 
    \textbf{Tasks:}
    \begin{itemize}
        \item Compare with standard features:
        \begin{itemize}
            \item Similar technology features
            \item Distinctive elements
        \end{itemize}
        \item Identify uniqueness deviations
        \item Remember errors.
    \end{itemize}

    \item \textbf{Final Scoring}  \\
    \textbf{Input:} Analysis results from previous steps \\ 
    \textbf{Output:} Final numerical score  \\
    \textbf{Patent Claim Validity \& Uniqueness Scoring (1-5 points):}
    \begin{itemize}
        \item 1 point (\textless 20\%):  
        \begin{quote}
        Example: "A data processing device comprising a processor and memory."  \\
        \textbf{Issues:} No technical features, generic components only, lack of differentiation, no uniqueness, vague implementation.
        \end{quote}

        \item 2 points (20-40\%):  
        \begin{quote}
        Example: "A data processing device with a high-speed processor and large memory." \\ 
        \textbf{Issues:} Minimal technical features, basic differentiation attempt, insufficient specificity, limited uniqueness, common technical level.
        \end{quote}

        \item 3 points (41-60\%):  
        \begin{quote}
        Example: "A data processing device comprising: a parallel processing processor; a high-speed cache memory; and a data optimization interface."  \\
        \textbf{Issues:} Basic technical features, some differentiation, basic uniqueness, partial specificity, some technical improvements.
        \end{quote}

        \item 4 points (61-80\%):  
        \begin{quote}
        Example: "A data processing device comprising: a multi-core parallel processor; a hierarchical cache memory structure; and a real-time data optimization interface."  \\
        \textbf{Strengths:} Clear technical features, specific differentiation, good uniqueness, concrete implementation, clear technical improvements.
        \end{quote}

        \item 5 points (\textgreater 80\%):  
        \begin{quote}
        Example: "A data processing device comprising: an AI-accelerated multi-core processor; a dynamic memory allocation hierarchical cache; and an intelligent interface with adaptive data optimization algorithms." \\ 
        \textbf{Strengths:} Strong technical features, excellent differentiation, high uniqueness, detailed implementation, innovative improvements.
        \end{quote}
    \end{itemize}
\end{enumerate}

\subsubsection{Ambiguous Claim Scope}

\subsection*{Name: Ambiguous Claim Scope}

\subsubsection*{I. Overview}
\textbf{Definition:} Evaluation of claim language clarity and boundary definition, ensuring specific and unambiguous protection scope without unclear or overly broad terms that could lead to uncertainty in claim interpretation.  

\textbf{Evaluation Purpose:} To ensure the claim clearly defines its protection boundaries using specific and precise language, avoiding ambiguous or overly broad terms that could weaken legal protection.

\subsubsection*{II. Key Evaluation Points}
\begin{enumerate}
    \item Boundary clarity verification
    \item Term specificity assessment
\end{enumerate}

\subsubsection*{III. Important Notes}
\begin{itemize}
    \item Check for clear scope boundaries
    \item Verify specific technical terms
    \item Confirm no ambiguous language
\end{itemize}

\subsubsection*{IV. Standard Examples}
\begin{quote}
"A data processing device, comprising:  
a 2.4 GHz processor specifically configured for image processing;  
a 512GB solid-state memory connected to the processor; and  
a 4K resolution display interface operating at 60Hz refresh rate."
\end{quote}

\subsubsection*{V. Evaluation Procedure}
\begin{enumerate}
    \item \textbf{Content Identification} \\ 
    \textbf{Input:} Claim 1 requiring evaluation  \\
    \textbf{Output:} Checklist, mapping, and scope classification  \\
    \textbf{Q:} Have all scope-defining elements been identified? \\ 
    \textbf{Tasks:}
    \begin{itemize}
        \item Carefully read the provided Claim 1 to identify:
        \begin{itemize}
            \item Scope boundaries
            \item Technical specifications
            \item Limiting terms
        \end{itemize}
        \item Verify term specificity:
        \begin{itemize}
            \item Technical parameters
            \item Functional limitations
        \end{itemize}
        \item Create mapping of scope elements
    \end{itemize}

    \item \textbf{Critical Element Verification} \\ 
    \textbf{Input:} Identified scope elements and boundaries \\ 
    \textbf{Output:} Scope verification report  \\
    \textbf{Q:} Do the elements clearly define boundaries?  \\
    \textbf{Tasks:}
    \begin{itemize}
        \item Confirm boundary clarity:
        \begin{itemize}
            \item Clear technical limits
            \item Specific definitions
        \end{itemize}
        \item Verify scope precision:
        \begin{itemize}
            \item Term specificity
        \end{itemize}
        \item Remember ambiguous or broad terms
    \end{itemize}

    \item \textbf{Format Compliance Assessment} \\ 
    \textbf{Input:} Scope verification results  \\
    \textbf{Output:} Compliance evaluation report  \\
    \textbf{Q:} Does the scope comply with clarity requirements?  \\
    \textbf{Tasks:}
    \begin{itemize}
        \item Check scope compliance:
        \begin{itemize}
            \item Term precision
            \item Boundary clarity
        \end{itemize}
        \item Remember clarity issues
    \end{itemize}

    \item \textbf{Standard-Based Comparative Analysis} \\ 
    \textbf{Input:} Compliance assessment results  \\
    \textbf{Output:} Comparative analysis report  \\
    \textbf{Q:} How well does it define protection scope?  \\
    \textbf{Tasks:}
    \begin{itemize}
        \item Compare with standard scope definitions:
        \begin{itemize}
            \item Similar technology scopes
            \item Accepted boundary definitions
        \end{itemize}
        \item Identify scope deviations
        \item Remember errors
    \end{itemize}

    \item \textbf{Final Scoring}  \\
    \textbf{Input:} Analysis results from previous steps \\ 
    \textbf{Output:} Final numerical score  \\
    \textbf{Patent Claim Scope Ambiguity Scoring (1-5 points):}
    \begin{itemize}
        \item 1 point (\textless 20\%):  
        \begin{quote}
        Example: "An improved device that processes data better." \\ 
        \textbf{Issues:} Highly ambiguous terms, unclear boundaries, unmeasurable improvements, subjective expressions, boundless protection scope.
        \end{quote}

        \item 2 points (20-40\%):  
        \begin{quote}
        Example: "An enhanced processing device with a high-speed processor and large memory."  \\
        \textbf{Issues:} Vague technical terms, broad scope, unclear parameters, generic expressions, wide protection scope.
        \end{quote}

        \item 3 points (41-60\%):  
        \begin{quote}
        Example: "A processing device comprising: a processor operating above 1GHz; and memory exceeding 256GB." \\ 
        \textbf{Issues:} Basic clarity, some specific parameters, basic boundary definition, measurable features, limited protection scope.
        \end{quote}

        \item 4 points (61-80\%):  
        \begin{quote}
        Example: "A processing device comprising: a 2.4GHz multi-core processor; a 512GB SSD; and a 4K resolution display." \\ 
        \textbf{Strengths:} Clear technical terms, specific parameters, clear boundaries, precise measurements, defined protection scope.
        \end{quote}

        \item 5 points (\textgreater 80\%):  
        \begin{quote}
        Example: "A processing device comprising: a 2.4GHz 8-core processor performing 1M operations/second; a 512GB NVMe SSD with 3000MB/s read speed; and a 4K 60Hz HDR display." \\ 
        \textbf{Strengths:} Very precise technical terms, clear performance parameters, exact boundary definitions, specific measurement criteria, precise protection scope.
        \end{quote}
    \end{itemize}
\end{enumerate}

\section{Prompts for generating by LLM}
\label{appendix:prompt_LLM}

You are an experienced patent attorney with a deep understanding of patent law and intellectual property (IP) protection. I will provide you with the title, abstract, and detailed description of a specific invention. Based on this information, please draft Claim 1 of the patent, ensuring that it accurately captures the essence of the invention while being legally strong and precise. Claim 1 should encompass the core technology of the invention while meeting legal requirements to distinguish it from prior art. It should use a clear and specific terminology to define the scope of protection appropriately, avoiding overly narrow or unnecessarily broad expressions.

Additionally, the structure of Claim 1 should adhere to the conventional format used in patent legal documents and incorporate terminology commonly used in the relevant technical field.

The Claim 1 you draft must meet the following criteria:

1. Clarity: Avoid ambiguous expressions and use precise language that can be easily understood.  
2. Legal Validity: Comply with patent examination standards and be legally appropriate to provide substantial protection.  
3. Breadth \& Differentiation: Adequately cover the core elements of the invention while highlighting its distinction from existing technology.  
4. Technical Accuracy: Accurately reflect the provided invention description without technical errors.  
5. Structural Integrity: Follow proper grammar and format for patent claims and maintain logical consistency.

Now, I will provide the patent title, abstract, and detailed description. Based on this information, please draft Claim 1 of the patent.

\end{document}